\documentclass[conference]{IEEEtran}
\IEEEoverridecommandlockouts
\usepackage{cite}
\usepackage{amsmath,amssymb,amsfonts}
\usepackage{algorithmic}
\usepackage{graphicx}
\usepackage{textcomp}
\usepackage{amssymb}
\usepackage{xcolor}
\usepackage{subfig}
\usepackage{url}
\def\BibTeX{{\rm B\kern-.05em{\sc i\kern-.025em b}\kern-.08em
    T\kern-.1667em\lower.7ex\hbox{E}\kern-.125emX}}
\begin{document}

\title{IGOT: Information Gain Optimized Tokenizer on Domain Adaptive Pretraining}

\author{\IEEEauthorblockN{Dawei Feng}
\IEEEauthorblockA{\textit{Tsinghua University} \\
Beijing, China \\
fdw22@mails.tsinghua.edu.cn}
\and
\IEEEauthorblockN{Yihai Zhang}
\IEEEauthorblockA{\textit{Tsinghua University} \\
Beijing, China \\
zhang-yh22@mails.tsinghua.edu.cn}
\and
\IEEEauthorblockN{Zhixuan Xu}
\IEEEauthorblockA{\textit{Tsinghua University} \\
Beijing, China \\
xzx21@mails.tsinghua.edu.cn}
}

\maketitle

\begin{abstract}
Pretrained Large Language Models (LLM) such as ChatGPT, Claude, etc. have demonstrated strong capabilities in various fields of natural language generation. However, there are still many problems when using LLM in specialized domain-specific fields. When using generative AI to process downstream tasks, a common approach is to add new knowledge (e.g., private domain knowledge, cutting-edge information) to a pretrained model through continued training or fine-tuning. However, whether there is a universal paradigm for domain adaptation training is still an open question. In this article, we proposed Information Gain Optimized Tokenizer (IGOT), which analyzes the special token set of downstream tasks, constructs a new subset using heuristic function $\phi$ with the special token and its information gain, to build new domain-specific tokenizer, and continues pretraining on the downstream task data. We explored the many positive effects of this method's customized tokenizer on domain-adaptive pretraining and verified this method can perform better than the ordinary method of just collecting data and fine-tuning. Based on our experiment, the continued pretraining process of IGOT with LLaMA-7B achieved 11.9\% token saving, 12.2\% training time saving, and 5.8\% maximum GPU VRAM usage saving, combined with the T5 model, we can even reach a 31.5\% of training time saving, making porting general generative AI to specific domains more effective than before. In domain-specific tasks, supervised $IGOT_\tau$ shows great performance on reducing both the convergence radius and convergence point during keep pretraining.
\end{abstract}

\begin{IEEEkeywords}
Large Language Model, Customized Tokenizer, Domain Adaption Pretraining,  Information Entropy
\end{IEEEkeywords}

\section{Introduction}

\begin{figure*}[htbp]
    \centering
    \includegraphics[width=1\linewidth]{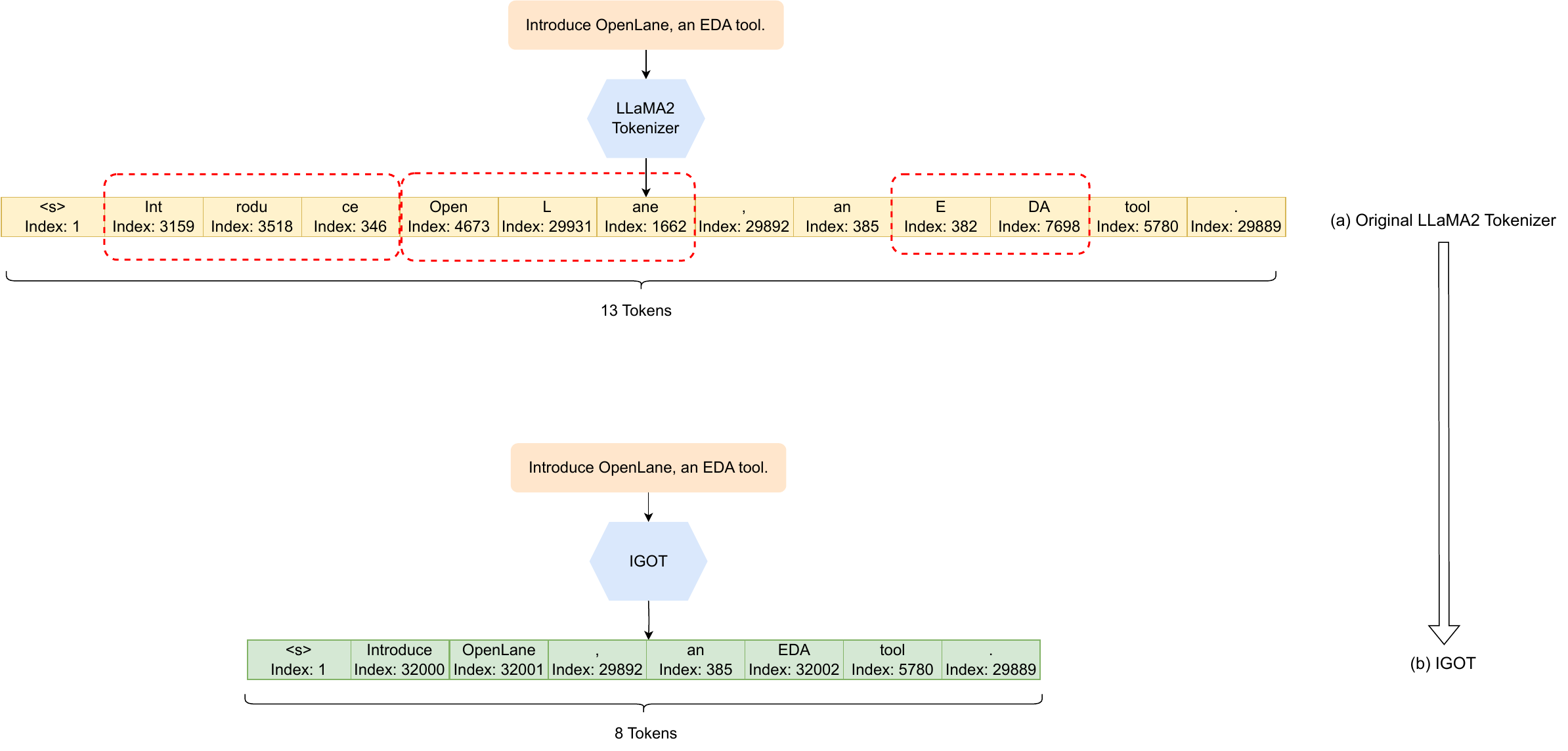}
    \caption{An example of IGOT compared to LLaMA2 Tokenizer. Given the input ``Introduce Openlane, an EDA tool'', the original LLaMA2 tokenizer will split into 13 tokens including meaningless parts such as ``rodu'' or ``ane'', as shown in Fig. 1(a). However, our IGOT method allows the tokenizer to process it into 8 meaningful parts, keep proprietary words like OpenLane, and reach a 38.46\% of space reduction.}
    \label{fig:tokenizer-openlane}
\end{figure*}

After extensive training on publicly available web data, large language models have exhibited remarkable capabilities in a wide range of common sense reasoning tasks\cite{brown2020language}. Some studies have indicated that these models, during pretraining, not only learn contextual text vector representations and probability distributions but are also likely to learn grammar\cite{vig2019multiscale}, syntax\cite{hewitt2019structural}, factual knowledge\cite{wang2020language}, even behavior patterns strikingly similar to human common sense\cite{davison2019commonsense}.

While general-purpose LLMs trained on massive internet datasets demonstrate impressive capabilities across diverse domains\cite{bubeck2023sparks}, recent work with domain-specific models like BloombergGPT \cite{wu2023bloomberggpt} and BioMedLM\cite{bolton2024biomedlm} highlights their superior performance within specific fields.

Early efforts to build large-scale language models utilized n-grams and simple smoothing techniques\cite{heafield2013scalable}\cite{buck2014n}. Various neural network architectures were later applied to the language modeling task, including feedforward networks\cite{bengio2000neural} and recurrent networks\cite{mikolov2010recurrent}. The introduction of the Transformer architecture\cite{vaswani2017attention} led to the development of highly scalable language models\cite{radford2019language}, demonstrating a predictable relationship between language modeling loss and scaling factors such as model size, number of training tokens, and compute budget.


The AI community has seen the release of numerous open-access LLMs, though they generally do not perform as well as their closed-access counterparts. ``Open-access LLM'' refers to models with publicly available weights, but transparency varies regarding training data and filtering techniques. EleutherAI released GPT-NeoX-20B\cite{black2022gpt} and GPT-J-6B \cite{wang2021codet5}, as well as the dataset these models were trained on Pile\cite{gao2020pile}. Google released UL2-20B\cite{tay2022ul2}, an encoder-decoder model trained on the publicly available C4\cite{raffel2020exploring}. Tsinghua University released the weights of GLM-130B\cite{zeng2022glm}, a Chinese-English LLM, and CodeGeeX-13B\cite{zheng2023codegeex}, a LLM for coding applications, without releasing the training sets. Salesforce released CodeGen-Mono-16B\cite{nijkamp2022codegen} without disclosing a proprietary Python dataset. Meta released the OPT\cite{zhang2022opt}, LLaMA\cite{touvron2023llama} under a non-commercial license and only provided high-level details about the data collection and filtering process.

Many attempts to fine-tune on specialized domain-specific datasets only use Supervised Fine-Tuning (SFT) or Keep Pretraining (KP), such as Lawyer LLaMA\cite{huang2023lawyer} or Huatuo\cite{wang2023huatuo}. However, after an in-depth analysis of the dataset and experiments using tokenizers of multiple pretrained models, we found that using pre-trained tokenizers on domain-specific datasets is very unwise. Tokenizers in general fields will produce redundancy several times or even dozens of times in the token space, resulting in higher training resource consumption and extended model convergence time.  By utilizing the IGOT approach to customize the tokenizer, our experiments demonstrate that the model achieves several-fold improvements in information representation. Additionally, this method conserves training resources and reduces fluctuations in the loss function during training, thereby enabling the model to converge to a more optimal point. To sum up, we make the following contributions:

\begin{itemize}
    \item Our IGOT method enhances the model's capability for information expression by analyzing downstream task data to identify the most effective specialized tokens for customizing the tokenizer. This process effectively transitions the model's tokenizer from a general domain to a vertical domain. Through mathematical methods and experiments, we analyze the reasons why IGOT enables the LLM to outperform traditional SFT or KP in multiple downstream tasks. This investigation elucidates the superior performance of IGOT across various applications.

    \item We show that the IGOT method can be a more universal approach for customizing models for downstream tasks. IGOT enhances model performance and conserves training resources by shifting the tokenizer's representational capabilities to specialized domains. From a mathematical perspective, we analyze why a customized tokenizer is essential for domain-specific LLMs, including information entropy and data compression considerations.
\end{itemize}

\section{Motivation}

Research by Chip-Chat\cite{blocklove2023chip} suggests that open-source LLMs fine-tuned on hardware design data (e.g., Verilog) outperform generic models for these tasks. This approach also avoids security risks associated with sending proprietary data to external LLMs. CyberRio\cite{cyberrio} also showed a way to use GPT-4 to harden a chip. 

During our experiments with LLMs in specialized domains, we observed unusual behavior in the vectorization of sentences containing domain knowledge by the original LLaMA2 tokenizer. For instance, as shown in Figure \ref{fig:tokenizer-openlane}(a), we continued pretraining the LLaMA2 using documentation from OpenLane\cite{9256623}, which is an automated RTL to GDSII flow based on components including OpenROAD, Yosys, Magic, Netgen, CVC, SPEF-Extractor, KLayout, and a number of custom scripts for design exploration and optimization. Subsequently, when we prompted the model to introduce OpenLane, we noted that the LLaMA2 tokenizer divided ``Introduce OpenLane, an EDA tool'' into 13 tokens, with ``Introduce'' being split into ``Int'', ``rodu'' and ``ce'' and ``OpenLane'' fragmented into ``Open'', ``L'', and ``ane''—splits that rendered the tokens meaningless.

As we mentioned in the introduction section, LLMs during their pretraining phase not only learn about contextual text vectorization and probability distributions but also acquire knowledge of grammar and syntax. The nonsensical splitting observed with the original tokenizer clearly results in a decreased ability of the subsequent Transformer architecture to capture information effectively. Consequently, we proposed our IGOT method, as illustrated in Figure \ref{fig:tokenizer-openlane}(b). This approach allows the tokenizer to process user inputs into more meaningful divisions during the tokenization stage, thereby enhancing the informational content of the sentences. This method ensures that the tokenizer adapts to domain-specific vocabularies and concepts, thus preserving the semantic integrity and enhancing the model's overall performance in specialized tasks.

\section{Method}
In this section, we will explain the implementation of our IGOT method, and its heuristic optimization $IGOT_\tau$. By maximizing information gain, the IGOT approach improves tokenization and enhances the model's ability to capture and utilize domain-specific knowledge more effectively.

\subsection{IGOT and $IGOT_\tau$}
In the context of classification systems, the information gain brought by a feature indicates its importance, with higher information gain suggesting greater significance. A basic formula for conditional entropy is given by:
\begin{equation}
    H(Y|X) = -\sum_{x \in X}\sum_{y \in Y} P(x,y) \log P(y|x)
\end{equation}

In the application of LLMs, our input undergoes tokenization by the tokenizer, resulting in a token list, which is then fed into the transformer architecture to generate probability distributions for subsequent vocabulary. Assuming the Transformer architecture can capture a context size of $\alpha$, with each token represented by an 8-bit integer (capable of representing up to 65536 different tokens), if a meaningful word $\delta$ is split into $N$ tokens, and a context consists of $\Delta$ words, we can define the unit of information obtained by the Transformer's window for this round as $\frac{N}{8\alpha}$. In an ideal scenario, our IGOT would process the word $\delta$ as one token. Thus, we can define the information gain $\theta$ of the idealized IGOT over the entire context as:

\begin{equation}
    \theta =  log(1+\sum_{i\in \Delta}N_{\delta_i})-log(1+\alpha)
\end{equation}

Under this premise, by setting a threshold $\epsilon$, we can simply construct a meaningful set of words $\lambda$ from the downstream task:

\begin{equation}
    \lambda = \{\theta_n>\epsilon, \ n\in \Delta \}
\end{equation}

However, using $\lambda$ selected directly based on information gain might not be the most efficient. Taking the OpenLane documentation as an example again, the $\lambda$ obtained by running the IGOT method contains proprietary names of chip IPs like ``sky130A\_sky130\_fd\_sc\_hd\_config'' which are typically very long and engineers often use a variety of custom symbols or abbreviations to distinguish their meanings. This causes such words to be split into many small tokens by the original tokenizer, resulting in very high information gain. Meanwhile, effective words that we actually hope to capture, such as ``Tritonroute'' may not achieve as high information gain, which is clearly not desirable. To address this issue, we introduce a new heuristic method $\phi$, which evaluates the information gain of words, their length, and the complete downstream vocabulary to produce a score. Based on this score, we then obtain a new subset of $\lambda$, denoted as $S_{\lambda}$:

\begin{equation}
    S_\lambda = \{ \phi(\theta_n;\ n;\ \Delta)> \epsilon^{'}, n \in \Delta \}
\end{equation}

We designate this heuristic-based method as $IGOT_\tau$.

\subsection{Heuristic $\phi$}

To obtain the optimal $\phi$ function, we use a simple neural network trained on a pre-annotated dataset D to develop an evaluator. This dataset D consists of common words and a large number of words like ``sky130A\_sky130\_fd\_sc\_hd\_config'' Each word is scored from 1 to 5 to indicate its desirability as a target word for our application. A simple idea is to use a multi-class optimization objective with cross-entropy as the loss function for optimization:

\begin{equation}
    \mathcal{L}(\Theta) = -\frac{1}{N} \sum_{i=1}^{N} \sum_{k=1}^{K} y_{ik} \log p_{ik} + \lambda \sum_{j} \theta_j^2
\end{equation}

However, we prefer our evaluator $\phi$ to provide scores rather than classifications. Therefore, we change our optimization objective to a regression task, aiming to minimize the L2 norm between predicted values and actual scores:

\begin{equation}
    \mathcal{L}(\Theta) = \frac{1}{N} \sum_{i=1}^{N} (y_i - f(x_i; \Theta))^2 + \lambda \sum_{j} \theta_j^2
\end{equation}

Training for multiple epochs with the Adam\cite{kingma2014adam} optimizer yields the $IGOT_\tau$ evaluator $\phi$.

\subsection{Domain Adaptive Pretraining}
In our investigation, we focused on the LLaMA2 model within the framework of the standard Causal Language Modeling (CLM) task. In this task, given an input token sequence $x=(x_0, x_1, x_2, \ldots)$, the model is trained to predict the next token $x_i$ in an autoregressive manner. Mathematically, the objective is to minimize the following negative log-likelihood:
\begin{equation}
    \mathcal{L}_{\textrm{CLM}} (\Theta) =\mathbb{E}_{x\sim\mathcal{D}_{\textrm{PT}}}\left[ -\sum_i\log p(x_i|x_0,\ldots,x_{i-1};\Theta)\right]
\end{equation}
Here, $\Theta$ represents the model parameters, $\mathcal{D}_{\textrm{PT}}$ is the pre-training dataset, $x_i$ is the token to be predicted, and $x_0,x_1,\ldots,x_{i-1}$ constitute the context.

To our Domain Adaptive Pretraining (DAP) target, we will mask part of the full task as the output, the loss is only calculated on the output part of the full sequence and can be expressed as:
\begin{equation}
    \mathcal{L}_{\textrm{DAP}} (\Theta) =\mathbb{E}_{x\sim\mathcal{D}_{\textrm{DAP}}}\left[ -\sum_{i\in \epsilon}\log p(x_i|x_0,\ldots,x_{i-1};\Theta)\right]
\end{equation}
Here, $\Theta$ represents the model parameters,  $\mathcal{D}_{\textrm{DAP}}$ is the Domain Adaptation Pretraining dataset, $\epsilon$ is the output sequence, $x=(x_0, x_1, \ldots)$ represents the tokenized input sequence.

\section{Experiment}
In Section 4.1, we begin by evaluating the impact of the IGOT method on the convergence of training loss and discussing the reasons behind this phenomenon. In Section 4.2, we generate heatmaps to analyze word distribution with information gain that results in effective information gain for specific datasets.

\subsection{IGOT for DAP}

\begin{figure}[htbp]
    \centering
    \subfloat[DAP with Original LLaMA2 Tokenizer]{\includegraphics[width=0.49\textwidth]{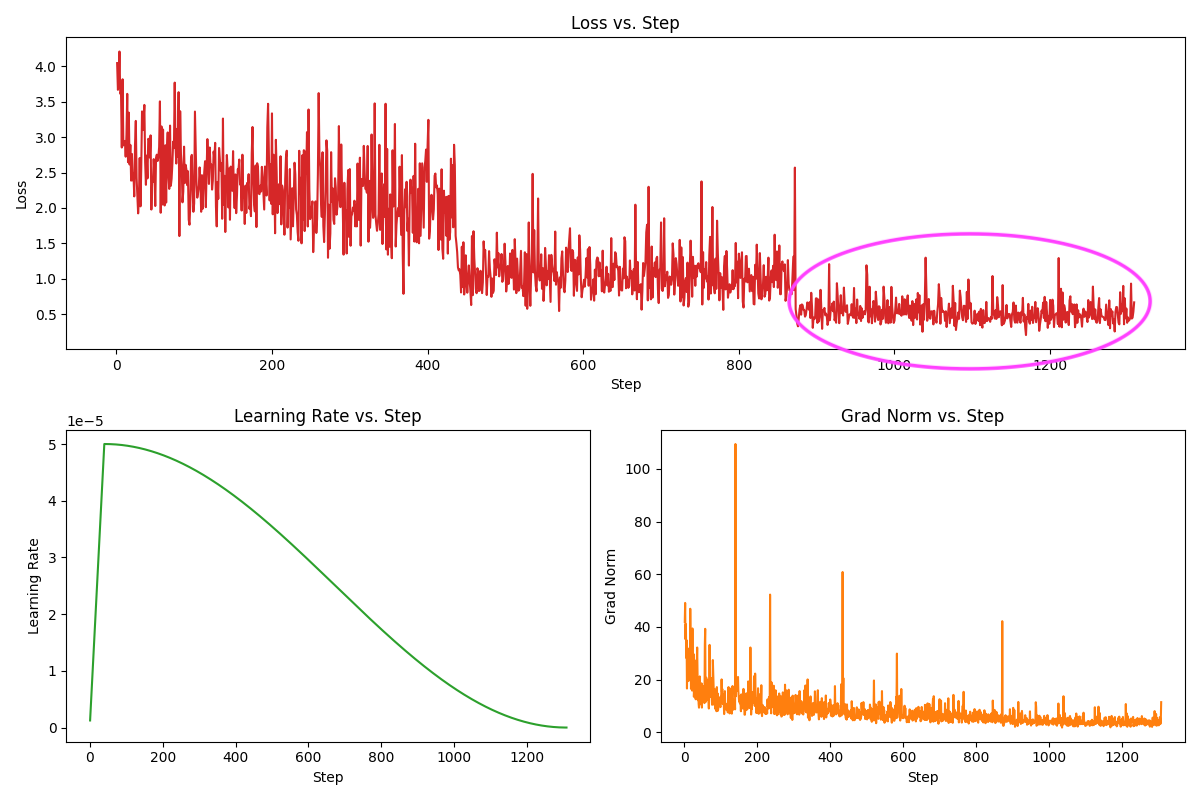}}\hfill
    \subfloat[DAP with IGOT]{\includegraphics[width=0.49\textwidth]{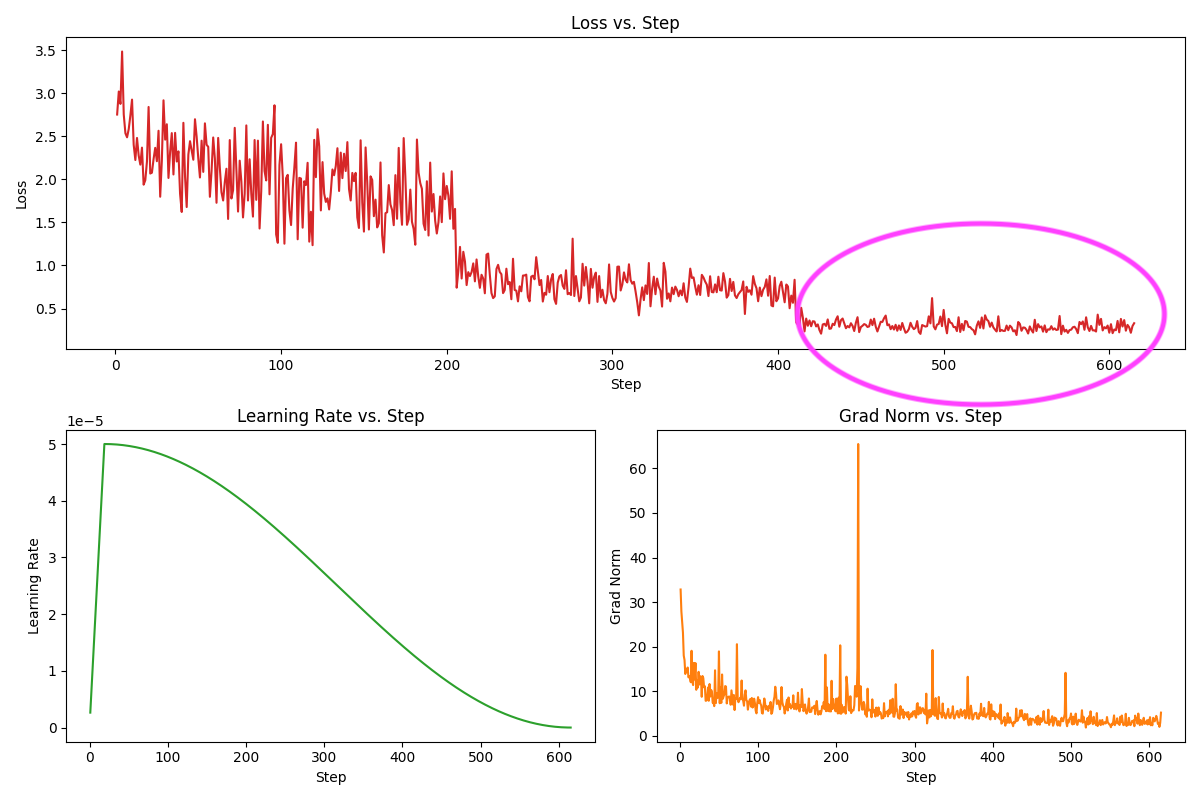}}
    \caption{In Figure (a), we showed the performance of the original LLaMA2-7B tokenizer during training. It is evident that even after the training reaches the third epoch, the loss oscillation remains severe. However, Figure (b) demonstrates that the IGOT method significantly reduces this oscillation, facilitating better model convergence.}
    \label{fig:IGOT-pretraining}
\end{figure}

We conducted the Domain Adaptive Pretraining (DAP) on LLaMA2-7B using OpenLane documentation and implemented the IGOT method. Figure \ref{fig:IGOT-pretraining} illustrates the impact of the IGOT method on the training loss convergence. Without the utilization of the IGOT method, the model's convergence radius remains relatively large, indicating substantial oscillation during training. This persistent oscillation can hinder the model's ability to converge to an optimal solution, leading to suboptimal performance on downstream tasks.

Based on the results presented in Table \ref{IGOT-cmp-util}, we achieved significant savings in computing resources, using IGOT with LLaMA2-7B on DAP leads to a 12.2\% saving in training time and 5\% savings in GPU VRAM. On the T5\cite{raffel2020exploring} series model, we were even able to achieve a 31.5\% time-saving. On Meta's OPT-350M model, there are also varying degrees of time and GPU VRAM savings. The efficacy of the IGOT method can be attributed to the customized tokenizer's ability to capture and incorporate domain-specific knowledge more effectively into the model's training process. By selecting optimal subsets of tokens based on their information gain, the IGOT method ensures that the tokenizer adapts to the nuances of the domain, resulting in more informative token representations. As a result, the model can better leverage domain-specific information, leading to improved performance on downstream tasks.

\begin{table}[htbp]
    \centering
    \caption{Computation savings brought by IGOT}
    \begin{tabular}{|l|lll|}
    \hline
        Approach &Added & Time Elapsed (s) & Max VRAM Usage\\ \hline
        LLama2-7B & 0 & 4799.43 & 154GB\\ 
        with $IGOT$& 5232 & 4210.57 (-12.27\%) & 146GB\\
        with $IGOT_{\tau}$& 640 & 4157.11 (-13.38\%)& 147GB\\ \hline
        OPT-350M& 0 & 86.34 & 22GB\\ 
        with $IGOT$& 3649 & 84.66 (-1.95\%)& 21GB\\
        with $IGOT_{\tau}$ & 619 & 80.09 (-7.23\%)& 21GB\\ \hline
        T5-Small& 0 & 292.88 & 14.5GB\\ 
        with $IGOT$& 4334 & 180.12 (-38.5\%)& 13GB\\
        with $IGOT_{\tau}$ & 615 & 241.45 (-17.56\%)& 13GB\\ \hline
        T5-Base& 0 & 301.16 & 28.7GB\\ 
        with $IGOT$& 4334 & 195.28 (-35.15\%)& 27GB\\
        with $IGOT_{\tau}$ & 615 & 251.02 (-16.64\%)& 27.5GB\\ \hline
    \end{tabular}
    \label{IGOT-cmp-util}
\end{table}

Furthermore, the use of a customized tokenizer also leads to more efficient training, as evidenced by the observed reduction in training time and resources. The more compact vectorization of sentences achieved by the customized tokenizer allows for faster computation and more efficient utilization of computational resources. This is particularly advantageous for large-scale model training, where even minor improvements in training efficiency can translate into significant time and cost savings.


\subsection{Token Distribution}

\begin{figure}[htbp]
    \centering
    \subfloat[$IGOT$]{\includegraphics[width=0.49\textwidth]{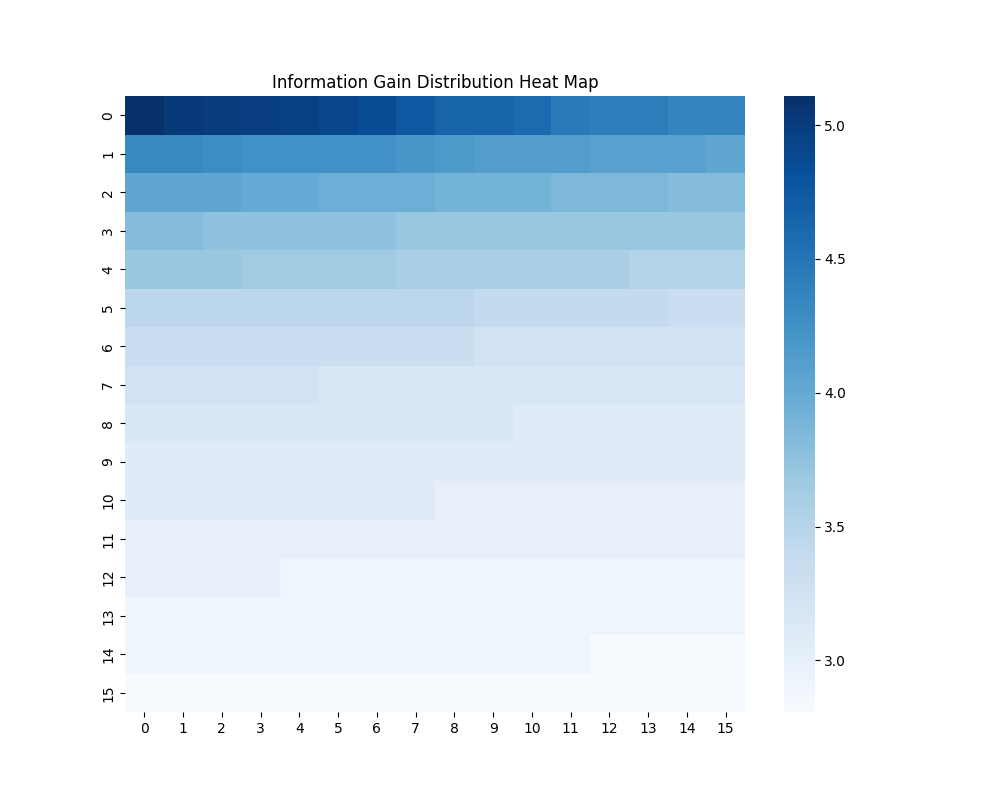}}\hfill
    \subfloat[$IGOT\tau$]{\includegraphics[width=0.49\textwidth]{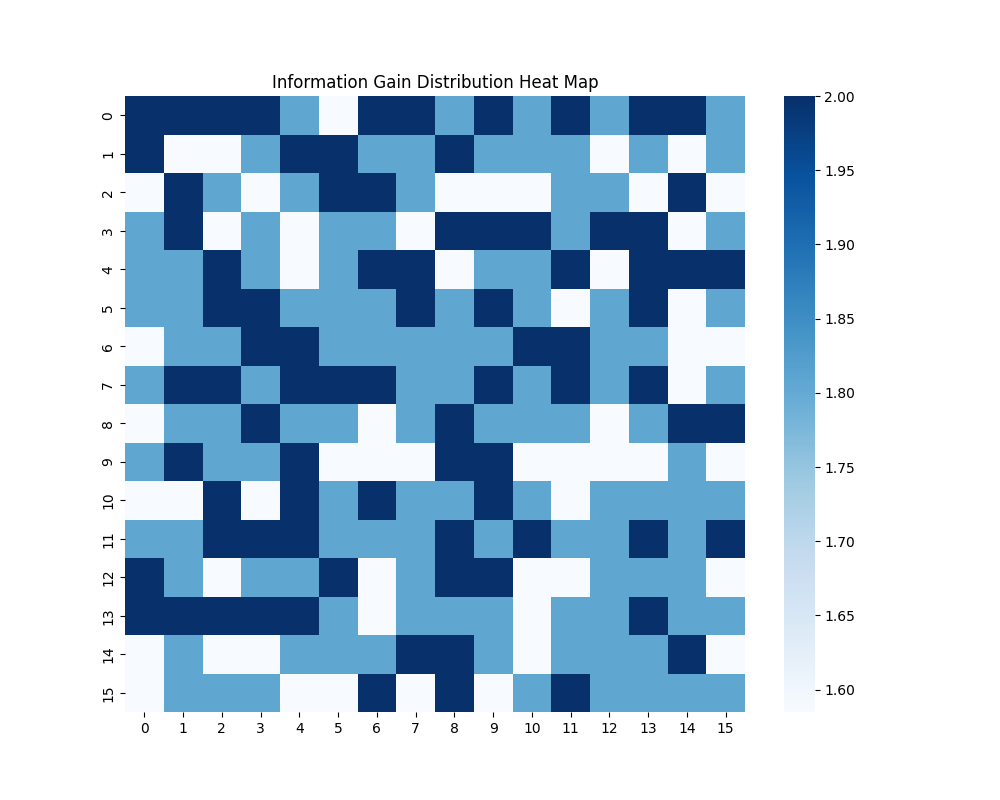}}
    \caption{Figure (a) depicts the distribution of information gain captured by IGOT on the complete dataset. Although many tokens with information gain greater than 4 are captured, the proportion is not high. Meanwhile, Figure (b) shows that the supervised $IGOT_\tau$ exhibits a more extensive distribution on high-gain tokens. This may be one of the reasons why the effectiveness of $IGOT_\tau$ is somewhat better than IGOT.}
    \label{fig:word}
\end{figure}

The analysis shown in Figure \ref{fig:word} provides insight regarding the distribution of tokens when using the IGOT method with OpenLane documentation. In Figure (a), we can see the distribution of information gain captured by IGOT across the entire dataset. While IGOT identifies many tokens with information gain higher than 4, the proportion of such tokens is relatively low. This suggests that although IGOT effectively identifies tokens with significant information gain, there is still room for improvement in capturing a higher proportion of high-gain tokens.

In contrast, Figure (b) illustrates the distribution of tokens when using supervised $IGOT_\tau$. Here, we see a more extensive distribution of high-gain tokens compared to the unsupervised IGOT method. This broader distribution indicates that supervised $IGOT_\tau$ is more capable of identifying and incorporating tokens with significant information gain into the tokenizer. Consequently, the enhanced coverage of high-gain tokens in supervised $IGOT_\tau$ may contribute to its superior effectiveness compared to the unsupervised IGOT method.

Future research could focus on refining the criteria used for token selection within the IGOT method. By developing more sophisticated algorithms or heuristics for identifying tokens with significant information gain, we can potentially enhance the effectiveness of the IGOT approach in capturing domain-specific knowledge more comprehensively.

Another area for future exploration is the investigation of alternative approaches to supervised token selection in the $IGOT_\tau$ variant. While manual selection of token subsets based on information gain can yield improvements, exploring automated or semi-automated methods for supervised token selection could further streamline the process and potentially lead to even greater gains in model performance.

Additionally, future work could involve the integration of domain-specific constraints or knowledge into the tokenization process. By incorporating domain-specific rules or guidelines into the tokenizer, we can ensure that the tokenization process aligns more closely with the linguistic characteristics and semantic nuances of the target domain, further enhancing the model's ability to capture and utilize domain-specific information effectively.

\section{Conclusion}




This study proposed and validated the Information Gain Optimized Tokenizer (IGOT) approach as a novel and effective strategy for adapting Large Language Models (LLMs) to specialized domains. Our IGOT method significantly enhances the domain-adaptation capabilities of LLMs by developing a customized tokenizer that optimally utilizes domain-specific token sets determined through heuristic function $\phi$ and their associated information gains.

Through our experimentation, including the use of LLaMA-7B for continued pretraining on specialized datasets such as OpenLane documentation, IGOT demonstrated a considerable improvement in model efficiency and effectiveness. Specifically, our findings showed both more than 10\% reduction in token usage and training time, with a 5.8\% reduction in maximum GPU VRAM usage. Such enhancements not only streamline the training process but also optimize computational resource utilization, making the adoption of general generative AI technologies in specialized domains more feasible and efficient.

Our work with IGOT also highlighted the capacity of this method to retain critical domain-specific information that traditional tokenization methods often overlook or mishandle. By enabling more meaningful and semantically coherent tokenization, IGOT not only enhances the quality of the model's output but also its ability to learn and apply new knowledge in domain-specific contexts more effectively.

\newpage
\bibliographystyle{ieeetr}
\bibliography{main.bib}


\end{document}